\documentclass[12pt]{article}
\usepackage[utf8]{inputenc}
\usepackage{graphicx}
\usepackage{float}  
\usepackage{amsmath}
\usepackage{booktabs}
\usepackage{placeins}  % Adds \FloatBarrier for better float control
\usepackage{plantuml}
\usepackage{geometry}  % Added for margin control
\usepackage{hyperref}  % Added for link handling

\title{Edge-AI for Agriculture: Lightweight Vision Models for Disease Detection in Resource-Limited Settings}
\author{Harsh Joshi \\ 
B.K. Birla College of Arts, Science and Commerce, Kalyan \\ 
\texttt{joshiharsh0506@gmail.com}}

\begin{document}
\maketitle
% Title for Abstract
\section*{Abstract}
% Placeholder Abstract Text
This research paper presents the development of a lightweight and efficient computer vision pipeline aimed at assisting farmers in detecting orange diseases using minimal resources. The proposed system integrates advanced object detection, classification, and segmentation models, optimized for deployment on edge devices, ensuring functionality in resource-limited environments. The study evaluates the performance of various state-of-the-art models, focusing on their accuracy, computational efficiency, and generalization capabilities. Notable findings include the Vision Transformer achieving 96 accuracy in orange species classification and the lightweight YOLOv8-S model demonstrating exceptional object detection performance with minimal computational overhead. The research highlights the potential of modern deep learning architectures to address critical agricultural challenges, emphasizing the importance of model complexity versus practical utility. Future work will explore expanding datasets, model compression techniques, and federated learning to enhance the applicability of these systems in diverse agricultural contexts, ultimately contributing to more sustainable farming practices.

% Title for Keywords
\section*{Keywords}
Computer Vision, Orange Disease Detection, Deep Learning, Edge Devices, Agricultural Technology, Model Optimization

% Introduction Section
\section{Introduction}
In the era of precision agriculture, the application of computer vision and deep learning technologies holds immense potential for automating critical tasks such as fruit classification and disease detection. These technologies promise to enhance crop management, improve yield prediction, and facilitate early disease intervention. However, the deploying such advanced systems in real-world agricultural settings faces significant challenges, primarily due to the scarcity of large, annotated datasets and the limited computational resources available to many farmers and agricultural workers. 

\medbreak

This study aims to develop lightweight and efficient computer vision pipelines that assist farmers in detecting orange diseases using minimal resources. The proposed system integrates object detection, classification, and segmentation models to provide an automated solution for detection, species classification and disease diagnosis. By optimizing these models for deployment on edge devices, the system is designed to function effectively even in resource-limited environments, making it accessible to a wider range of users. Furthermore, the study evaluates the performance of several state-of-the-art models, examining their accuracy, computational efficiency, and generalization capabilities. Through rigorous testing and validation, the study demonstrates the potential of computer vision in addressing real-world agricultural challenges and improving disease management for orange crops. 

\section{Literature Review}
Recent advancements in computer vision have significantly impacted agricultural image processing, particularly in the domains of semantic segmentation, disease detection, and crop classification. Convolutional Neural Networks (CNNs) have emerged as powerful tools for these tasks, with architectures like U-Net and Fully Convolutional Networks (FCNs) demonstrating superior performance \cite{Lei2024-th} . achieved 98\% accuracy in fruit and vegetable classification using DenseNet with squeeze-excitation blocks, highlighting the potential of enhanced CNN architectures \cite{inbook}. However, the scarcity of high-quality labeled datasets often hinders model performance, necessitating the development of various data augmentation techniques, including Manipulation-Based Data Augmentation (MBDA) and synthetic image generation using Generative Adversarial Networks (GANs) \cite{Nitin2023-jj},\cite{Shorten2019-bl}.
\medbreak
Disease detection in agriculture has seen notable progress through the integration of multiple data sources and advanced architectures. Chen et al. developed a framework combining image and sensor data, achieving 94\% mIoU in lesion segmentation, surpassing traditional models like UNet++ \cite{Chen2024-kg}. This trend towards data fusion and specialized architectures represents a significant advancement in agricultural disease detection capabilities. In citrus disease detection, CNNs have significantly improved identification accuracy for diseases such as Greening, Canker, Anthracnose, and Melanose \cite{Qiu2023-rw}. The integration of image-text multimodal fusion and knowledge assistance has shown promise in improving detection accuracy in complex backgrounds, suggesting a potential direction for future work \cite{Qiu2023-rw}.

\medbreak
The field has evolved to incorporate multiclass semantic segmentation for comprehensive agricultural analysis. Khan et al. introduced a multifaceted approach integrating frequency-domain image co-registration and transformer-based segmentation, achieving over 94\% accuracy in crop classification \cite{Khan2024-gw}. Li et al. enhanced agricultural image segmentation through an Agricultural Segment Anything Model Adapter (ASA), significantly improving segmentation accuracy for specific agricultural tasks \cite{Li2023-uu}. Despite these advancements, challenges persist in developing robust models for agricultural applications. Future research should explore advanced data augmentation techniques, integrate multimodal data, and develop scalable, unified frameworks for comprehensive agricultural analysis [1,\cite{Ngugi2024-yg}. The ongoing development of specialized models and techniques promises to further enhance the capabilities of computer vision systems in agriculture, potentially leading to more efficient and accurate crop management and disease detection methods.

\medbreak

This review addresses key challenges in agricultural computer vision: creating integrated, scalable systems and implementing multi-task learning for field deployment. We propose a cohesive framework combining orange detection, counting, species classification, and disease segmentation. Using oranges as our test subject, we aim to develop a resource-efficient, scalable pipeline. Our approach seeks to optimize real-time processing and enhance generalizability across diverse orange varieties and growing conditions. 

% Research Methodology Section
\section{Research Methodology}
The oranges were sourced from multiple markets, representing a diverse array of orchards, which introduces variability across different environmental and cultivation practices. This diversity enhances the dataset's representativeness, supporting model generalization to varied real-world conditions. Mainly 5 species of Oranges (Tangerine, Navel, Blood Oranges, Bergmout, Tangelo) and Disease species (Citrus cranker, Black spots, Sooty mould, Blue-green mould, Citrus greening) were chosen. For creating the dataset, the images were annotated with bounding boxes to created object detection dataset and then exported into coco and yolo format. The classification task specific dataset with 5 orange species (47-52 images per class ) was created by taking images of the oranges from various angles, under various lighting condition and a smart phone was used to imitate real world scenarios. In the case of segmentations data, it was created using same images where disease were annotated as per the class using review of the farmer.

\medbreak

\begin{figure}[H]
\centering
\includegraphics[width=0.8\textwidth]{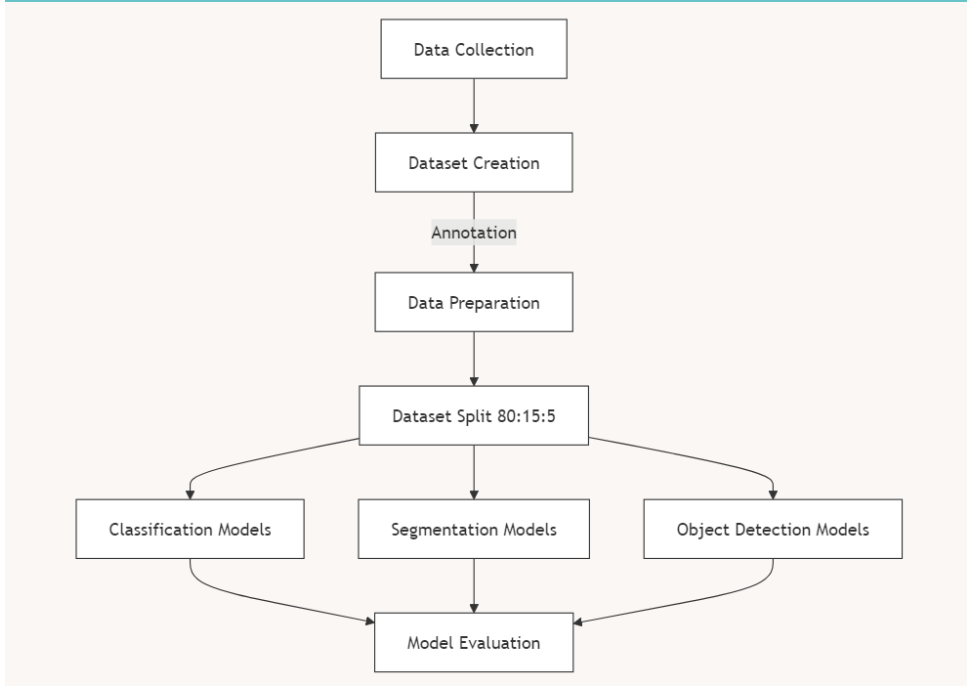}
\caption{Research Methodology}
\label{fig:reserch methodology}
\end{figure}

Methodology of this paper (Figure: \ref{fig:reserch methodology} ), For image classification a set of algorithms was evaluated, the algorithms consist of convolutional models (ResNet50 \cite{He2015-eg}, EfficientNet \cite{Tan2019-kq}, InceptionNet \cite{Szegedy2014-jl}) and transformer-based \cite{Vaswani2017-xb} architectures (Vision Transformer (ViT) \cite{Dosovitskiy2020-mp}, coatnet \cite{Dai2021-ep}, Shifted window Transformer (swin)\cite{Liu2021-kk}), regnet \cite{Xu2021-ft}, DenseNet \cite{8099726}, and MobileNetV3 \cite{Howard2019-yp}. Additionally knowledge distillation \cite{Hinton2015-va} (ResNet+ DenseNet) and Teacher-Student model were used, to improve performance by transferring learned features from larger to more efficient models.  Each model was initialized with ImageNet \cite{5206848} pretrained weights to facilitate transfer learning and improve convergence .The TIMM Python package was used for training, and models were fine-tuned using Cross Entropy Loss with the Stochastic Gradient Descent (SGD) optimizer, combined with a ReduceLROnPlateau learning rate scheduler over 50 epochs. Early stopping and learning rate decay were implemented to adaptively control learning rates and prevent overfitting \cite{Goodfellow-et-al-2016}. Model's performance was evaluated on basis of Recall, Precision, F1-score, Accuracy and Area under RoC curve and cross validation (5-fold) was applied to comprehensively capture model performance across various classification aspects \cite{Powers2020-ak}. 

For the task of multi class semantic segmentation in detecting diseases in oranges, several models were employed, including DeepLabV3 \cite{Chen2017-pb}, Feature Pyramid Network (FPN) \cite{8099589}, LinkNet \cite{Chaurasia2017-rb}, U-Net++ \cite{10.1007/978-3-030-00889-5_1}, Detectron2 with Mask R-CNN \cite{noauthor_undated-ne}, and You Only Look Once V8 (YOLOv8) Segmentation \cite{jocher2023yolo},\cite{Chiu2020-yv}, \cite{7780460}. These models were selected for their capability to achieve high pixel-level precision in agricultural image segmentation. The training configuration included a Softmax2D activation function, CrossEntropyLoss, and the Adam optimizer \cite{Kingma2014-dp} with weight decay, along with a ReduceLROnPlateau scheduler for adaptive learning rate adjustment over 50 epochs \cite{Everingham2010-qb}. Model performance was evaluated based on Intersection over Union (IoU), F1-Score, Precision, Recall, Pixel Accuracy, and Model Size to assess segmentation accuracy and computational efficiency \cite{Powers2020-ak}.

\medbreak

To perform object detection for identifying and localizing oranges within images, various algorithms were employed, including YOLO [29,30,31], RetinaNet \cite{Lin2017-fz}, DETR \cite{Carion2020-sh}, along with their respective versions. These models were trained on over 600 labeled instances, enabling accurate detection of individual oranges, even within cluttered or complex backgrounds. The bounding boxes generated from this detection phase were subsequently utilized in classification and segmentation stages to facilitate more detailed analysis. Model performance was evaluated using metrics such as mean Average Precision (mAP), Precision, Recall, F1-Score, Intersection over Union (IoU), and Model Size, to assess detection accuracy and computational efficiency 

\medbreak

The dataset was split in ratio of 80:15:5 (train:test:validation) while training. The nvidia graphics card of 6 gb vram and 8gb ddr4 ram was used while training. 

\begin{figure}[h!]
\centering
\includegraphics[width=0.8\textwidth]{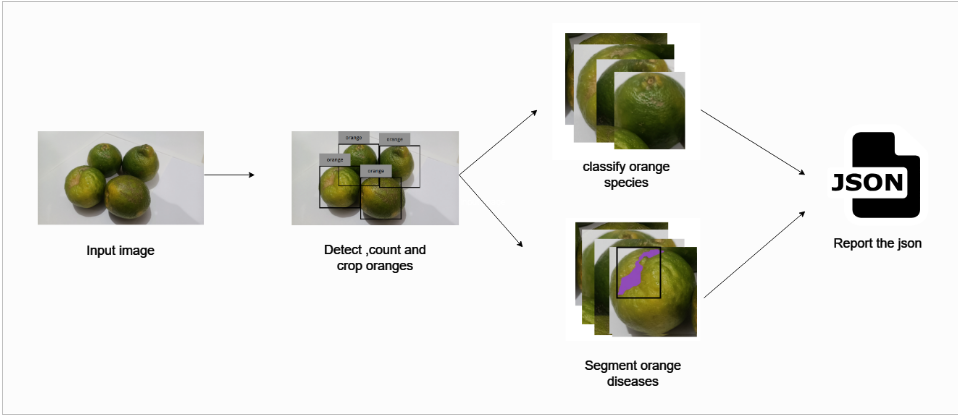}
\caption{Figure 1 : Working of API for Oranges }
\label{fig:work flow}
\end{figure}

For deployment purposes, the top-performing models for classification, object detection, and segmentation were exported. The API operates in two main sections: computation and reporting. Initially, the input image undergoes object detection, where oranges are detected, counted, and cropped based on bounding box coordinates. These cropped images are then processed in parallel by the classification and segmentation models, allowing for more focused and accurate analysis.

\medbreak

The classified and segmented images are then reassembled, overlaying bounding boxes on the original image to display the identified orange species and any segmented disease regions. In addition to the final annotated image, the API generates a JSON file containing detailed information, including processing time for each step, confidence scores for each predicted class, bounding box coordinates, and the file paths for cropped images (Figure \ref{fig:work flow} ). This structured output supports API evaluation based on latency, accuracy, response time, and throughput, offering valuable insights for agricultural applications \cite{KOIRALA2019219}. The combination of detection, classification, and segmentation within a unified pipeline, inspired by architectures like Mask R-CNN \cite{8237584}, enables real-time analysis and reporting, facilitating timely decision-making in agricultural practices and fruit disease detection \cite{HADIPOURROKNI2023106611}. 

% Results Section
\section{Results}
\textbf{Classification results}

The study compared the performance of several deep learning models, ranging from lightweight architectures like MobileNetV3 to complex ensembles such as ResNet + DenseNet, using weighted ROC AUC, accuracy, and macro-averaged precision, recall, and F1-score. Significant performance variations were observed, reflecting differences in model architecture and their ability to handle multi-class classification of orange species. 

\begin{table}[h!]
\centering
\renewcommand{\arraystretch}{1.2} % Increase row height
\caption{Classification models evaluation}
\adjustbox{max width=\textwidth}{ % Scale table dynamically
\begin{tabular}{@{}lccccc@{}}
\toprule
\textbf{Model}                  & \textbf{Accuracy} & \textbf{Precision} & \textbf{Recall} & \textbf{F1 Score} & \textbf{Size (MB)} \\ \midrule
DenseNet 201                    & 0.94              & 0.95               & 0.95            & 0.95              & 70.1               \\
NF-RegNet-B1                    & 0.92              & 0.93               & 0.93            & 0.93              & 35.44              \\
ResNet 152D                     & 0.93              & 0.94               & 0.94            & 0.93              & 222.74             \\
Swin-Tiny-Patch4-Window7-224    & 0.91              & 0.92               & 0.92            & 0.92              & 105.05             \\
CoAtNet Nano RW 224             & 0.94              & 0.95               & 0.95            & 0.95              & 55.95              \\
EfficientNet B5                 & 0.91              & 0.92               & 0.91            & 0.91              & 109.05             \\
MobileNet V3                    & 0.93              & 0.95               & 0.92            & 0.93              & 24                 \\
Vision Transformer (ViT)        & 0.96              & 0.95               & 0.95            & 0.95              & 323                \\
Inception Next Small            & 0.94              & 0.95               & 0.95            & 0.95              & 179.89             \\
ResNet + DenseNet               & 0.94              & 0.93               & 0.94            & 0.92              & 292.75             \\
SE-ResNet-50                    & 0.93              & 0.93               & 0.93            & 0.93              & -                  \\ \bottomrule
\end{tabular}}
\label{table:classification model evaluation}
\end{table}

Table \ref{table:classification model evaluation} presents comprehensive performance metrics for various classification models, including accuracy, precision, recall, and F1-score. Vision Transformer (ViT) emerged as the top performer, achieving exceptional performance with 96\% accuracy and 0.95 F1-score across all metrics. However, its larger size of 323MB suggests significant computational requirements.
\medbreak
CoAtNet Nano stands out as the second-best performer, matching DenseNet 201's metrics (0.94 accuracy, 0.95 F1-score) while being notably more efficient at just 55.95MB. This makes it an excellent balance of performance and resource efficiency. The architectural contrast between ViT's pure transformer approach and CoAtNet's hybrid design combining convolution and attention mechanisms highlights the evolution in model architectures.
\medbreak
MobileNet V3 emerges as the most impressive lightweight solution, achieving 0.93 accuracy and 0.93 F1-score while requiring only 24MB of storage. This exceptional performance-to-size ratio makes it ideal for resource-constrained deployments. While not matching ViT's peak performance, MobileNet V3's ability to maintain competitive metrics while being 13x smaller demonstrates the effectiveness of mobile-optimized architectures.

\begin{table}[h!]
\centering
\renewcommand{\arraystretch}{1.2}
\caption{ Resource Utilization Metrics }
\adjustbox{max width=\textwidth}{%
\begin{tabular}{@{}p{4cm}p{2.5cm}p{5cm}p{5cm}@{}}
\toprule
\textbf{Model} & \textbf{Memory Usage} & \textbf{Resource Efficiency} & \textbf{Deployment Feasibility} \\ \midrule
DenseNet 201 & Moderate & High accuracy with moderate resource use & Feasible for balanced resource settings \\
NF-RegNet B1 & Low & Lightweight with good efficiency & Suitable for low-resource environments \\
ResNet 152D & High & Robust accuracy with high memory demand & Only for high-resource settings \\
Swin Tiny Patch4 224 & Moderate & Efficient model with moderate size & Suitable for balanced use cases \\
CoAtNet Nano RW 224 & Low & Optimal balance of accuracy and size & Ideal for constrained resource settings \\
EfficientNet B5 & Moderate & Modest size and resource requirements & Suitable for mixed environments \\
MobileNet V3 & Very Low & Excellent portability and efficiency & Ideal for edge devices \\
Vision Transformer (ViT) & Very High & Highest accuracy with high resource use & Suitable for high-performance settings \\
Inception Next Small & High & High accuracy with high resource needs & Feasible in well-equipped environments \\
ResNet + DenseNet & Very High & Strong performance but high resource use & Restricted to high-resource availability \\ \bottomrule
\end{tabular}%
}
\label{table:Resource Utilization Metrics}
\end{table}

\begin{figure}[h!]
\centering
\includegraphics[width=0.6\textwidth]{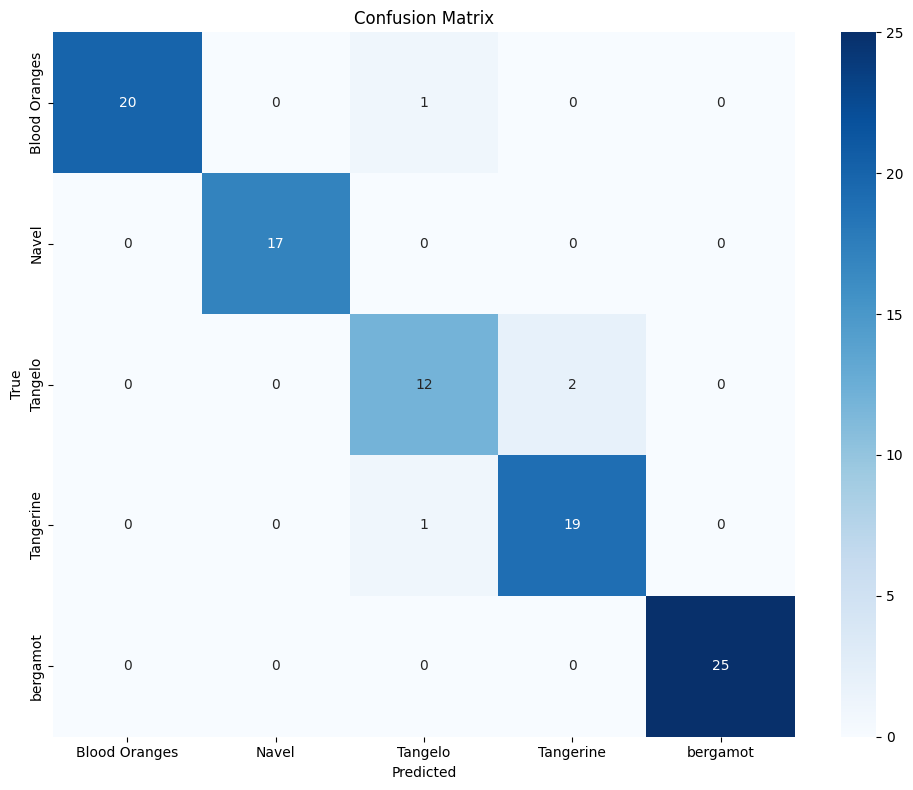}
\caption{Confusion matrix VIT}
\label{fig:Confusion matrix VIT}
\end{figure}

As illustrated in the confusion matrix (Figure \ref{fig:Confusion matrix VIT}), the ViT model demonstrated excellent classification performance across all five orange species. Notably, it achieved perfect classification for Blood Oranges, Navel, and Bergamot. Minor misclassifications were observed for Tangelo and Tangerine, with 2 Tangelo samples misclassified as Tangerine and 1 Tangerine sample misclassified as Tangelo. 

\begin{figure}[h!]
\centering
\includegraphics[width=0.6\textwidth]{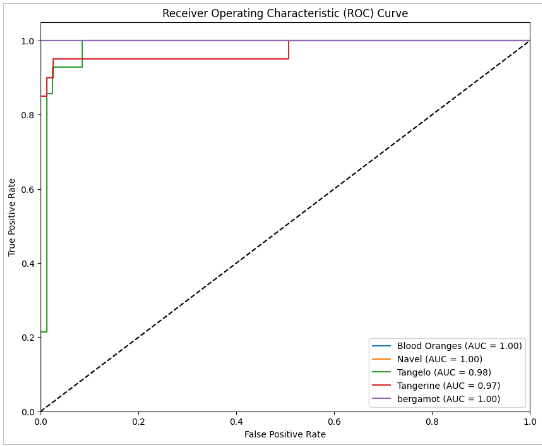}
\caption{RoC Curve for VIT}
\label{fig:RoC Curve for VIT}
\end{figure}

The Area Under the Curve (AUC) for ViT was 0.98 (Figure \ref{fig:RoC Curve for VIT}), reinforcing its robustness in handling multi-class classification. The comprehensive classification evaluation report confirmed near-perfect precision and recall scores across all species, emphasizing the model’s ability to capture long-range dependencies in image data. This makes ViT particularly effective at distinguishing subtle differences between orange species. 

\medbreak

Given the multi-class nature of the task, macro-averaged metrics provide a comprehensive view of model performance across all classes, preventing skewness due to class imbalance. Weighted ROC AUC values across models reaffirm the reliability of these models in distinguishing between orange species, with ViT and ResNet-50 setting new benchmarks in classification accuracy. 

\medbreak

\textbf{2. Segmentation Model Performance}

For segmentation tasks, the models were evaluated based on Intersection over Union (IoU), pixel accuracy, precision, recall, and F1-score. These tasks focused on identifying diseases such as Blackspot, Sooty Mold, and Citrus Canker on orange surfaces. 

\begin{table}[h!]
\centering
\renewcommand{\arraystretch}{1.2}
\caption{ Segmentations models evaluation }
\adjustbox{max width=\textwidth}{%
\begin{tabular}{@{}lcccccc@{}}
\toprule
\textbf{Model}   & \textbf{IoU}   & \textbf{F1-Score} & \textbf{Precision} & \textbf{Recall} & \textbf{Pixel Accuracy} & \textbf{Model Size} \\ \midrule
DeepLab          & 0.8983         & 0.9206            & 0.9077             & 0.9392          & 0.9392                  & 85.7 MB             \\
FPN              & 0.8987         & 0.9187            & 0.8987             & 0.9432          & 0.9432                  & 22 MB               \\
LinkNet          & 0.9039         & 0.9255            & 0.9196             & 0.9435          & 0.9435                  & 270 MB              \\
U-Net++          & 0.8967         & 0.9222            & 0.9186             & 0.9319          & 0.9319                  & 140 MB              \\ \bottomrule
\end{tabular}%
}
\label{table:Segmentations models evaluation}
\end{table}

Among the tested models, LinkNet demonstrated the highest IoU (0.9039) and F1-score (0.9255), shown in table \ref{table:Segmentations models evaluation} outperforming the others in segmenting diseased regions. DeepLab followed closely with comparable metrics, while FPN achieved similar accuracy despite its significantly smaller size (22 MB). U-Net++, while effective, had a slightly lower IoU (0.8967) but maintained strong performance in terms of precision and recall.

\medbreak

While larger models such as LinkNet and DeepLab achieved the highest segmentation performance, their substantial size may restrict their applicability in real-world scenarios. In contrast, lightweight models like FPN provide a viable alternative, offering competitive performance with a significantly reduced model size. This trade-off between model size and accuracy is crucial for applications in resource-constrained environments. 
\medbreak

\textbf{3. Object detection}

\begin{table}[ht!]
\centering
\renewcommand{\arraystretch}{1.4}
\caption{ Object detection models evaluation }
\adjustbox{max width=\textwidth}{%
\begin{tabular}{@{}p{3cm}p{2.5cm}p{2.5cm}p{2.5cm}p{2.5cm}p{2.5cm}@{}}
\toprule
\textbf{Metric}                      & \textbf{YOLO v8 s} & \textbf{RetinaNet + ResNet50} & \textbf{RetinaNet + ResNet101} & \textbf{DETR (66MB)} & \textbf{DETR (Better)} \\ \midrule
\textbf{mAP50}                       & 0.949              & 0.933                         & 0.9127                         & 0.892                & 0.9461                 \\
\textbf{mAP50-95}                    & 0.600              & 0.872                         & 0.8623                         & 0.512                & 0.5955                 \\
\textbf{mAP75}                       & -                  & 0.913                         & 0.9013                         & -                    & -                      \\
\textbf{Precision}                   & 0.884              & -                             & -                              & 0.859                & 0.9170                 \\
\textbf{Recall}                      & 0.921              & -                             & 0.895 (maxDets=10)             & 0.785                & 0.8548                 \\
\textbf{F1 Score (max)}              & 0.910              & 0.910                         & 0.8789                         & 0.820                & 0.8848                 \\
\textbf{Optimal Confidence Threshold} & 0.402              & 0.402                         & -                              & -                    & -                      \\
\textbf{Inference Speed}             & 10.9 ms/image      & -                             & -                              & 17.3 ms/image        & -                      \\
\textbf{mAP (medium objects)}        & -                  & 0.626                         & 0.6045                         & -                    & -                      \\
\textbf{mAP (large objects)}         & -                  & 0.898                         & 0.8926                         & -                    & -                      \\
\textbf{Model Size}                  & 6 MB               & -                             & -                              & 66 MB                & -                      \\ \bottomrule
\end{tabular}%
}
\label{table: Object detection models evaluation}
\end{table}

YOLOv8-S stands out for its exceptional speed (10.9ms per image) and accuracy (mAP50: 0.949), making it ideal for real-time applications. It achieves a high balance between precision and recall (Table \ref{table: Object detection models evaluation}).
\medbreak
RetinaNet with ResNet 50 is the second-best, offering strong mAP50 (0.933) and excellent performance on both medium and large objects, while balancing precision, recall, and F1 score effectively. 

\begin{figure}[H]
\centering
\includegraphics[width=0.6\textwidth]{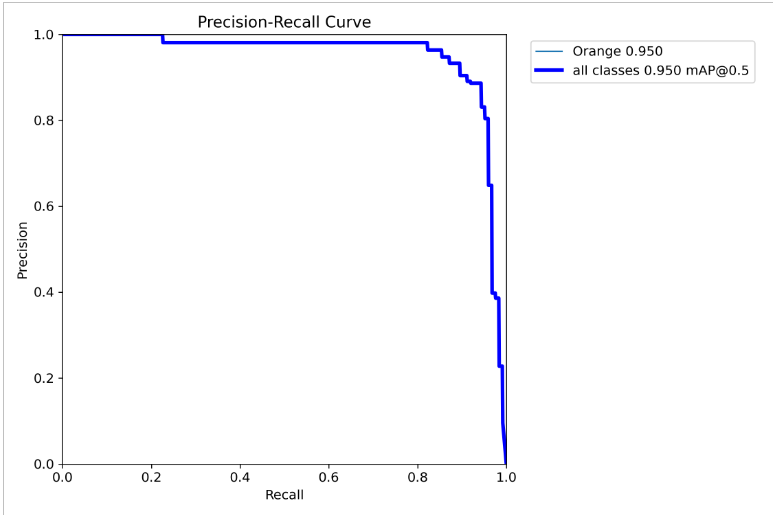}
\caption{ Precision-Recall Curve for YOLO-8s}
\label{fig:PR curve}
\end{figure}

Figure \ref{fig:PR curve} shows The curve maintains high precision (near 1.0) for most recall values, dropping sharply only at very high recall. This suggests excellent performance, with a mean Average Precision (mAP) of 0.950 at 0.5 IoU threshold for all classes.

\begin{figure}[H]
\centering
\includegraphics[width=0.6\textwidth]{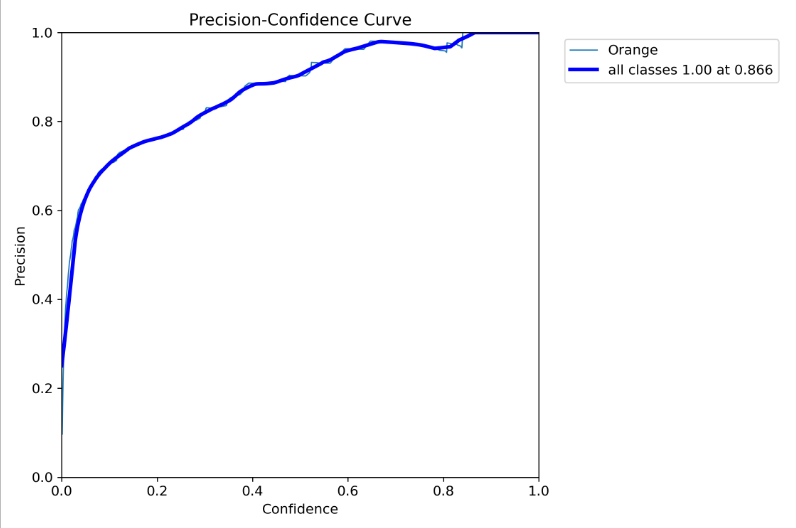 }
\caption{Precision-Confidence Curve for YOLO-8s}
\label{fig:Pc curve}
\end{figure}

The curve in figure \ref{fig:Pc curve} shows high precision across most confidence levels, reaching 1.0 at 0.866 confidence. It indicates the model maintains good precision even at lower confidence thresholds, with a sharp initial rise and gradual improvement thereafter.

\begin{figure}[H]
\centering
\includegraphics[width=0.6\textwidth]{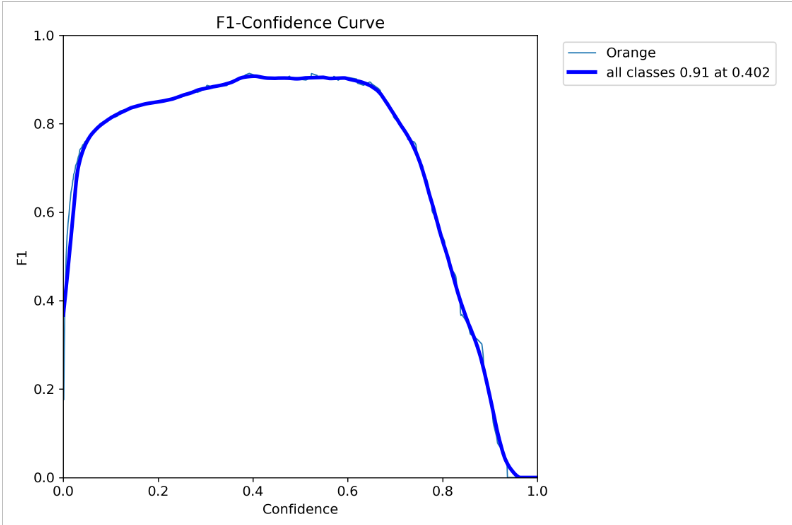 }
\caption{Precision-Confidence Curve for YOLO-8s}
\label{fig:f1 conf curve}
\end{figure}

The F1 score curve figure \ref{fig:f1 conf curve},  peaks at 0.91 at a confidence of 0.402, then remains stable before declining at higher confidences. This indicates a good balance between precision and recall, with optimal performance at moderate confidence levels.

% Discussion Section
\section{Discussion}
Our research addresses key challenges in agriculture, particularly for small-scale farmers in resource-constrained environments. Notably, our selected models achieve high accuracy with approximately 50 training images per class, demonstrating efficient learning from limited datasets. This efficiency is especially beneficial in agricultural contexts, where large datasets are difficult to obtain due to resource limitations and crop-specific variations \cite{8253600}. Achieving robust performance with small datasets accelerates deployment in new settings and supports adaptability across different crop varieties, making it a valuable tool for precision agriculture.
Our findings highlight the potential of lightweight computer vision pipelines for classifying orange species and detecting diseases—a topic of significant interest in recent agricultural research [4]. Among the models evaluated, the Vision Transformer (ViT) provided the highest accuracy for classification, while LinkNet outperformed others in segmentation tasks, underscoring the value of advanced architectures in precision agriculture \cite{KAMILARIS201870},42]. However, the trade-off between accuracy and computational efficiency remains an important consideration (Table \ref{table:Resource Utilization Metrics} ), especially when deploying models in low-resource environments.
\medbreak
To address such challenges, we prioritized lightweight models for deployment. YOLOv8-S (6 MB) for object detection, MobileNet (24 MB) for classification, and FPN (22 MB) for segmentation emerged as the best candidates due to their small size and suitability for edge devices. This choice is critical for farmers in rural areas or developing regions, where high-performance infrastructure may be lacking \cite{9869273}. Optimizing model size and computational demand expands access to advanced agricultural technology, making it feasible for small-scale farmers to adopt these solutions without expensive hardware.\cite{SINGH2024124387}
\medbreak
An additional advantage of our pipeline is its speed, processing images in approximately 1 second per instance while maintaining high accuracy. Real-time processing is essential in agricultural applications, as rapid decision-making can significantly affect crop health and quality. For example, a conveyor belt in a packing facility could efficiently sort oranges by species and detect potential diseases, enhancing quality control processes \cite{BHARGAVA2021243}. This balance between speed and accuracy ensures the practicality of our pipeline for real-world agricultural use.
\medbreak
Our approach demonstrates high accuracy with limited data, making it especially suited to specialized agricultural applications where extensive datasets are often unavailable \cite{article_1339516}. This efficiency with small datasets also facilitates deployment in diverse environments and adaptability across crop varieties, addressing a significant need in precision agriculture \cite{10.1007/978-3-030-35990-4_12}. Furthermore, deploying lightweight models, such as YOLOv8-S for object detection, allows for effective usage on edge devices or in low-resource areas, expanding accessibility for small-scale farmers. Achieving this balance between accuracy and computational demands helps optimize resource utilization, making advanced agricultural solutions viable for those with limited infrastructure.
\medbreak
Despite performing well with a small dataset, our models would benefit from further testing on larger, more diverse datasets to enhance robustness under various environmental conditions. Expanding the dataset to include a broader range of lighting, growth stages, and rare orange species could improve model generalizability and application across different agricultural contexts. Although Vision Transformer (ViT) showed strong performance, its size (323 MB) may limit use in highly resource-constrained settings. Future work could focus on model compression techniques to reduce its size without sacrificing accuracy, enabling high-performing models on low-power devices.
\medbreak
As this study may not encompass the full range of environmental conditions or orange species, a broader dataset incorporating factors like variable weather, soil types, and growth stages could further increase applicability. Future research could also explore federated learning to leverage distributed datasets while ensuring data privacy. Incorporating temporal data might enable disease progression tracking, while developing models that adapt to local conditions with minimal retraining could enhance practical utility in precision agriculture.

% Conclusion Section
\section{Conclusion}
This research demonstrates the efficacy of modern deep learning architectures in addressing critical agricultural challenges through computer vision. The Vision Transformer achieved remarkable performance in orange species classification with 96\% accuracy and a 0.95 F1-score, while LinkNet demonstrated superior segmentation capabilities with an IoU of 0.9039. The lightweight YOLOv8-S model's exceptional performance in object detection (mAP50: 0.949) with minimal computational overhead (10.9 ms/image) particularly stands out. These results establish that sophisticated computer vision tasks can be effectively accomplished with limited training data, a crucial finding for specialized agricultural applications.
\medbreak
The study reveals a nuanced relationship between model complexity and practical utility. While larger models like ViT (323 MB) achieved superior accuracy, lightweight alternatives such as MobileNet (24 MB) and FPN (22 MB) demonstrated competitive performance with significantly reduced computational demands. This trade-off becomes particularly significant in resource-constrained agricultural environments, where the ability to process images in approximately one second while maintaining high accuracy represents a viable solution for real-time applications.
\medbreak
Our findings have broader implications for the democratization of agricultural technology. The success of these models with just 50 training images suggests a practical pathway for developing specialized agricultural applications where large datasets are traditionally unavailable. However, future research should address limitations in environmental variability coverage and explore federated learning approaches for leveraging distributed datasets. Additionally, investigating model compression techniques and temporal data analysis could further enhance the practical utility of these systems. These advancements could significantly impact small-scale farming operations by providing accessible, efficient tools for crop monitoring and disease detection, ultimately contributing to more sustainable agricultural practices. 

% References Section
\section { References }

\bibliography{ref.bib}

\begin{thebibliography}{10}
\bibitem{noauthor_undated-ne}
detectron2: Detectron2 is a platform for object detection, segmentation and other visual recognition tasks.
\end{thebibliography}
\bibliographystyle{unsrt}

% \printbibliography
\end{document}